\documentclass{article}

\usepackage[preprint,nonatbib]{neurips_2026}

\usepackage[utf8]{inputenc}
\usepackage[T1]{fontenc}
\usepackage[hidelinks]{hyperref}
\usepackage{url}
\usepackage{graphicx}
\usepackage{booktabs}
\usepackage{amsfonts}
\usepackage{nicefrac}
\usepackage{microtype}
\usepackage{xcolor}
\usepackage{amsmath}
\usepackage{float}

\title{Mind the Unseen Mass: Unmasking LLM Hallucinations via Soft-Hybrid Alphabet Estimation}

\author{%
  Hongxing Pan \quad Yingying Guo \quad Wenqing Kuang \quad Jiashi Lu \\
  School of Data Science \\
  The Chinese University of Hong Kong, Shenzhen \\
  \texttt{\{124090486, 123090142, 123090247, 123090386\}@link.cuhk.edu.cn} \\
}

\begin{document}

\maketitle

\begin{abstract}
This paper studies uncertainty quantification for large language models (LLMs) under black-box access, where only a small number of responses can be sampled for each query. In this setting, estimating the effective semantic alphabet size---that is, the number of distinct meanings expressed in the sampled responses---provides a useful proxy for downstream risk. However, frequency-based estimators tend to undercount rare semantic modes when the sample size is small, while graph-spectral quantities alone are not designed to estimate semantic occupancy accurately. To address this issue, we propose SHADE (Soft-Hybrid Alphabet Dynamic Estimator), a simple and interpretable estimator that combines Generalized Good--Turing coverage with a heat-kernel trace of the normalized Laplacian constructed from an entailment-weighted graph over sampled responses. The estimated coverage adaptively determines the fusion rule: under high coverage, SHADE uses a convex combination of the two signals, while under low coverage it applies a LogSumExp fusion to emphasize missing or weakly observed semantic modes. A finite-sample correction is then introduced to stabilize the resulting cardinality estimate before converting it into a coverage-adjusted semantic entropy score. Experiments on pooled semantic alphabet-size estimation against large-sample references and on QA incorrectness detection show that SHADE achieves the strongest improvements in the most sample-limited regime, while the performance gap narrows as the number of samples increases. These results suggest that hybrid semantic occupancy estimation is particularly beneficial when black-box uncertainty quantification must operate under tight sampling budgets.
\end{abstract}

\section{Introduction}

Large language models can hallucinate or contradict reliable evidence \cite{ji2023survey,rawte2023survey,Farquhar2024DetectingHI}. Reliable uncertainty quantification (UQ) supports abstention and human oversight in risk-sensitive settings \cite{Shorinwa2024ASO,Liu2025UncertaintyQA}. In deployed systems, however, one often faces a \emph{strict budget}: only a few independent generations per query are affordable for monitoring, and proprietary APIs hide logits, activations, and token probabilities \cite{chen2024quantifying,Xue2025VerifyWU,DBLP:conf/aaai/SunLHCJLHH26}. This paper targets that \textbf{small-$n$ black-box} regime.

Semantic \emph{alphabet size}---the number of semantic equivalence classes obtained by clustering multiple generations under bidirectional entailment---is an interpretable proxy for how ``spread out'' model meanings are for a query \cite{DBLP:conf/iclr/KuhnGF23,Farquhar2024DetectingHI,mccabe2025estimating}. At very small $n$, both purely frequency-based and purely spectral estimates tend to underestimate effective support: empirical counts ignore geometric organization among draws, while eigenvalues of a semantic graph can be unstable unless anchored to occupancy statistics.

\begin{figure}[h]
  \centering
  \includegraphics[width=0.85\linewidth]{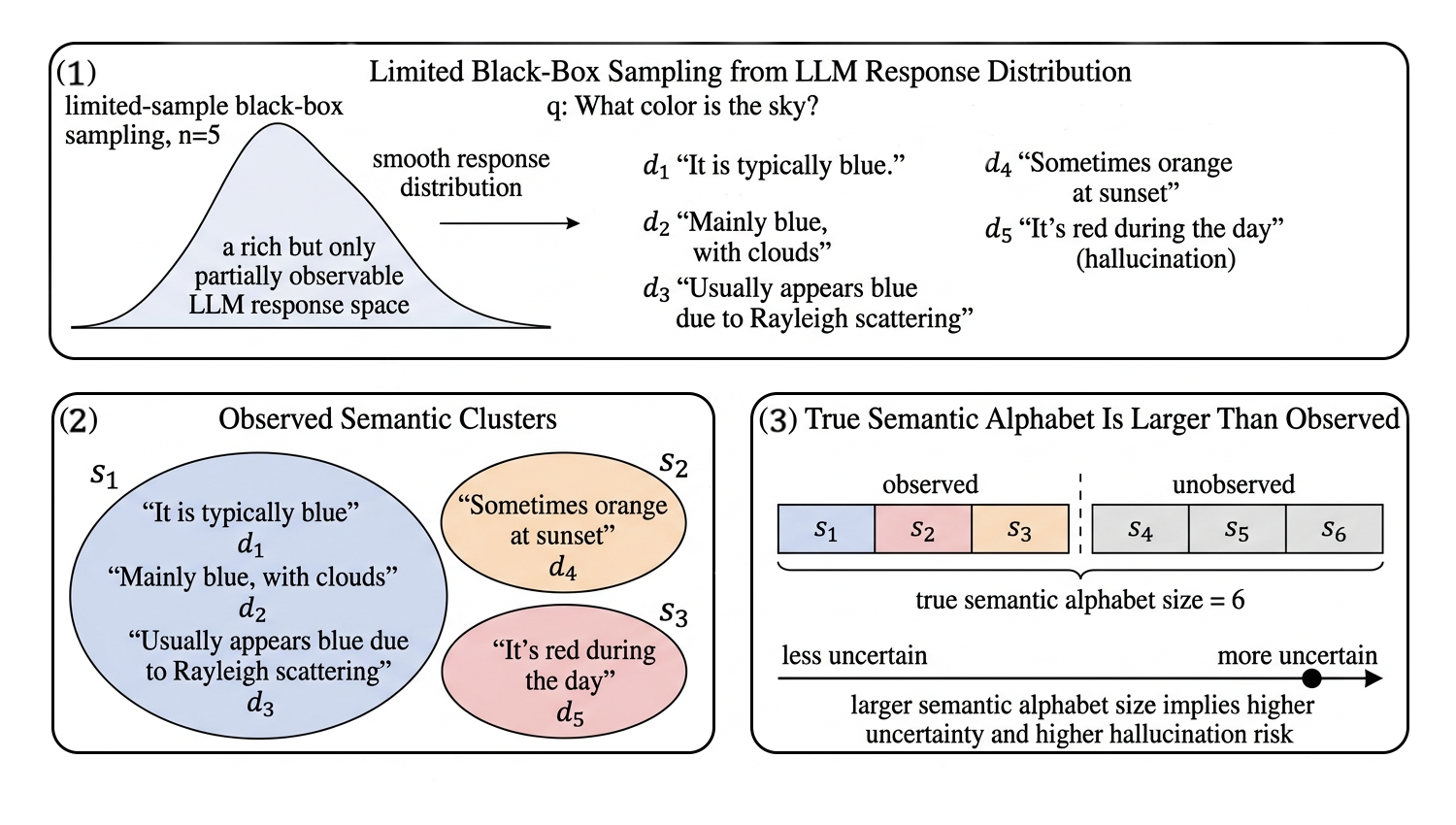}
  \caption{Black-box sampling observes only a few responses per query; clustering yields $k_{\text{obs}}$ semantic classes, while the \emph{true} semantic alphabet can be larger due to missing mass. Larger effective alphabet implies higher epistemic uncertainty and higher risk of inconsistent outputs under the same monitoring budget.}
  \label{fig:semantic_budget}
\end{figure}

We propose \textbf{SHADE} (Soft-Hybrid Alphabet Dynamic Estimator). SHADE combines (i) missing-mass extrapolation via Generalized Good--Turing (GGT) and (ii) a diffusion-based spectral summary, the heat-kernel trace $\mathrm{tr}(e^{-\beta L})$ of the normalized Laplacian of an entailment-weighted graph over the $n$ responses \cite{chung1997spectral}. Estimated coverage $C_{\text{GGT}}$ determines \emph{how} the two signals are fused: a convex combination when coverage is high, and a LogSumExp surrogate when coverage is low. A lightweight finite-sample correction stabilizes the hybrid cardinality before a Horvitz--Thompson--style entropy readout used as a risk score \cite{chao2003nonparametric}. Compared with occupancy-only cardinality \cite{mccabe2025estimating} and graph-density features for UQ \cite{Li2025EnhancingUQ}, the graph here serves as a \emph{second estimator of the same scalar}, fused through coverage rather than as an auxiliary feature vector.

\paragraph{Contributions.}
\begin{enumerate}
  \item A \textbf{coverage-gated hybrid} between GGT-based mass extrapolation and the heat-kernel trace of a semantic graph, avoiding a hard threshold on $n$ alone.
  \item A \textbf{single pipeline} from raw generations to a bias-corrected cardinality and a visibility-adjusted entropy suitable for thresholding.
  \item \textbf{Empirical analysis} of alphabet-size error and downstream incorrectness detection: gains concentrate at the smallest sampling budgets.
\end{enumerate}

\section{Preliminary}

\paragraph{Entropy and semantic classes.}
For a discrete variable over classes $s$ with probabilities $p(s)$, Shannon entropy is $\mathbb{H}=-\sum_s p(s)\log p(s)$. Semantic entropy (SE) partitions generations into equivalence classes using bidirectional entailment \cite{DBLP:conf/iclr/KuhnGF23,Farquhar2024DetectingHI,Kossen2024SemanticEP}. With white-box access, class probabilities can be integrated from token likelihoods; in the black-box setting, one uses empirical class frequencies $\hat{p}_i$, yielding discrete semantic entropy (DSE) that plugin-evaluates entropy but \emph{underestimates} diversity when heavy tails leave most classes unseen \cite{chen2024quantifying,Farquhar2024DetectingHI}.

\paragraph{Coverage and graphs.}
Let $f_m$ denote the number of semantic classes that appear exactly $m$ times in $n$ samples. The missing mass $M$ of unobserved classes and coverage $C=1-M$ are central to Good--Turing style reasoning \cite{good1953population}. Independently, $n$ responses induce a weighted undirected graph: nodes are responses, edges carry entailment-based weights, and the normalized Laplacian $L$ encodes global connectivity \cite{Li2025EnhancingUQ,Nikitin2024KernelLE}. Eigenvalues $\lambda_i$ of $L$ describe how strongly responses split into modes versus clump together \cite{chung1997spectral}.

\section{Related work}
\label{sec:related}

\paragraph{Hallucination detection and UQ for LLMs.}
Extracting robust signals from noisy or limited observations is a ubiquitous challenge spanning spatial modeling, medical diagnostics, and representation learning \cite{cui2025efficient, tang2025bilateral, zeng2026learning, cui2024superpixel}. In the context of generative AI, a large literature studies this problem through the lens of hallucinations and uncertainty in LLMs \cite{ji2023survey,rawte2023survey,Shorinwa2024ASO,Liu2025UncertaintyQA}. Practical UQ methods include self-consistency, evidential models, internal probes, and semantic clustering approaches \cite{chen2024quantifying,Chen2024INSIDELI,yoon2025uncertainty,Kossen2024SemanticEP}. Our focus is the intersection of \emph{black-box} access and \emph{small} $n$.

\paragraph{Semantic entropy and structure.}
Semantic Uncertainty and Semantic Entropy establish clustering-by-meaning as a paradigm \cite{DBLP:conf/iclr/KuhnGF23,Farquhar2024DetectingHI}. Follow-up work uses pairwise similarity, kernelized structure, evidential objectives, and adaptive exploration \cite{nguyen2025beyond,Nikitin2024KernelLE,kunitomo2026evidential,DBLP:conf/aaai/SunLHCJLHH26}. McCabe et al.~\cite{mccabe2025estimating} study semantic cardinality from occupancy; Li et al.~\cite{Li2025EnhancingUQ} incorporate graph density as an auxiliary UQ signal. SHADE differs: the Laplacian spectrum contributes a \emph{parallel} estimate of effective support fused with GGT through $C_{\text{GGT}}$.

\paragraph{Graph spectra and estimation.}
Graph representations of multi-sample generations appear in several lines of work \cite{Derr2018SignedGC,Huang2022AreGC,Feng2020GraphRN,bui2025effectiveness}. Unlike training-heavy graph models, SHADE uses the graph only as a structural estimator combined with classical missing-mass statistics, keeping inference lightweight.

\section{Methodology}
\label{sec:method}

Let $k_{\text{obs}}$ be observed semantic classes after clustering $n$ generations. We estimate effective support by combining a mass-based $\hat{|S|}_{\text{GGT}}$ and a spectral $\hat{|S|}_{\text{Soft-EigV}}$. 
\paragraph{Heat-kernel trace.}
Build symmetric weights $w_{ij}=(a_{ij}+a_{ji})/2$ from NLI entailment probabilities $a_{ij}$ \cite{DBLP:journals/corr/abs-2111-09543}. Let $L$ be the normalized Laplacian with eigenvalues $0=\lambda_1\le\cdots\le\lambda_n\le 2$ \cite{chung1997spectral}. The heat kernel $e^{-\beta L}$ diffuses mass on the graph; its trace
\begin{equation}
\hat{|S|}_{\text{Soft-EigV}} := \mathrm{tr}(e^{-\beta L}) = \sum_{i=1}^{n} e^{-\beta \lambda_i}
\end{equation}
aggregates low-frequency (coherent) structure with exponential damping on high-frequency noise \cite{chung1997spectral}. Thus $\mathrm{tr}(e^{-\beta L})$ acts as a \emph{soft} multiscale count of semantic modes complementary to raw $k_{\text{obs}}$.

\paragraph{GGT coverage.}
Let $f_1,f_2$ be singleton and doubleton class counts. We estimate missing mass and coverage as in stabilized GGT formulations \cite{mccabe2025estimating,good1953population}:
\begin{equation}
M_{\text{GGT}} = \frac{1}{n} \left(1 - \frac{2.08}{n^{0.7}}\right) f_1 + \frac{4.1}{n^{1.7}} f_2,\quad
C_{\text{GGT}} = \max(1 - M_{\text{GGT}}, 10^{-12}),\quad
\hat{|S|}_{\text{GGT}} = \frac{k_{\text{obs}}}{C_{\text{GGT}}}.
\end{equation}

\paragraph{Coverage-driven hybridization.}
When $C_{\text{GGT}}\ge \tau$, we use a convex combination that down-weights the spectrum as coverage grows:
\begin{equation}
\hat{|S|}_{\text{Hybrid}} = C_{\text{GGT}}\, \hat{|S|}_{\text{GGT}} + (1-C_{\text{GGT}})\, \hat{|S|}_{\text{Soft-EigV}}.
\end{equation}
When $C_{\text{GGT}}<\tau$, we use a LogSumExp fusion that behaves like a smooth maximum between the two predictors:
\begin{equation}
\hat{|S|}_{\text{Hybrid}} = \frac{1}{\alpha} \ln \left( e^{\alpha \hat{|S|}_{\text{GGT}}} + e^{\alpha \hat{|S|}_{\text{Soft-EigV}}} \right).
\end{equation}
Scalars $(\beta,\alpha,\tau)$ are fixed once on development data (Sec.~\ref{sec:experiments}). The threshold $\tau$ is chosen so typical queries avoid unstable switching near the boundary.

\paragraph{Finite-sample correction and entropy readout.}
Plugin diversity functionals incur $\mathcal{O}(1/n)$ bias \cite{miller1955bias}; we subtract a leading term of the same order from the hybrid cardinality:
\begin{equation}
\hat{|S|}_{\text{Final}} = \hat{|S|}_{\text{Hybrid}} + \frac{k_{\text{obs}} - 1}{2n},\qquad
p_i^* = \frac{k_{\text{obs}}\,\hat{p}_i}{\hat{|S|}_{\text{Final}}}.
\end{equation}
The score used for detection is
\begin{equation}
\hat{\mathbb{H}}_{\text{SHADE}} = -\sum_{i=1}^{k_{\text{obs}}} \frac{p_i^* \log p_i^*}{1 - (1 - p_i^*)^n},
\end{equation}
with a visibility denominator standard under sampling without replacement \cite{chao2003nonparametric}. Ablations $\widehat{|S|}_{\mathrm{Hybrid}}$ and $\hat{\mathbb{H}}_{\mathrm{Hybrid}}$ omit this correction before the entropy mapping.

\section{Experiments}
\label{sec:experiments}

\subsection{Setup}

We evaluate on SQuAD, CoQA, NQ-Open, TriviaQA, and HotpotQA \cite{DBLP:conf/emnlp/RajpurkarZLL16,DBLP:journals/tacl/ReddyCM19,DBLP:journals/tacl/KwiatkowskiPRCP19,DBLP:data/12/JoshiCWZ25,DBLP:conf/emnlp/Yang0ZBCSM18}. Generators include OPT-6.7B, Qwen3-8B-Instruct\footnote{\url{https://huggingface.co/Qwen/Qwen3-8B}}, Mistral-7B-Instruct, and Phi-3.5-mini \cite{DBLP:journals/corr/abs-2205-01068,DBLP:journals/corr/abs-2310-06825,DBLP:journals/corr/abs-2404-14219}. DeBERTa-v3-large-mnli supplies entailment scores for graph construction \cite{DBLP:journals/corr/abs-2111-09543}. For alphabet-size error, we draw $N{=}100$ generations per query as a pseudo-oracle and subsample $n\in\{5,\dots,50\}$; baselines include plugin $k_{\text{obs}}$, GT, GGT, Laplacian $U_{\text{EigV}}$, and hybrid variants. Binary incorrectness labels follow dataset protocols ($AUC_s$, $AUC_r$); CoQA participates in estimation pools but is omitted from the four-dataset AUC table for space. Reproducibility details are in Appendix~\ref{app:exp}.

\subsection{Alphabet-size estimation}

Table~\ref{tab:mae_results} reports MAE and RMSE against the $N{=}100$ reference. SHADE achieves the largest margin at $n{=}5$ and remains best across listed $n$. MAE is not strictly monotone in $n$ for any method because pooling mixes heterogeneous prompts. Table~\ref{tab:pairwise_results} summarizes pairwise win rates: SHADE beats prior hybrids and frequency baselines on a majority of queries.

\begin{table}[h]
\centering
\caption{MAE (RMSE) vs.\ $N{=}100$ oracle across subsample sizes $n$.}
\label{tab:mae_results}
\small
\begin{tabular}{lccccc}
\toprule
Estimator & $n{=}5$ & $n{=}8$ & $n{=}10$ & $n{=}25$ & $n{=}50$ \\
\midrule
Plugin ($k_{\mathrm{obs}}$) & 4.18 (6.34) & 3.83 (5.87) & 3.65 (5.60) & 2.66 (4.24) & 1.54 (2.62) \\
Good-Turing (GT)           & 3.66 (5.55) & 3.44 (5.12) & 3.30 (4.86) & 2.50 (3.85) & 1.43 (2.32) \\
Generalized GT (GGT)       & 3.60 (5.57) & 3.43 (5.22) & 3.30 (4.96) & 2.48 (3.81) & 1.42 (2.30) \\
U-EigV (Laplacian)         & 3.98 (6.16) & 3.66 (5.73) & 3.51 (5.49) & 2.88 (4.59) & 2.47 (3.93) \\
Hybrid (GT+EigV)           & 3.52 (5.52) & 3.22 (4.99) & 3.06 (4.70) & 2.28 (3.63) & 1.38 (2.23) \\
Hybrid-GGT                 & 3.40 (5.42) & 3.20 (5.03) & 3.06 (4.76) & 2.27 (3.60) & 1.38 (2.21) \\
\midrule
\textbf{SHADE (Ours)}      & \textbf{2.86 (4.82)} & \textbf{2.83 (4.37)} & \textbf{3.01 (4.25)} & \textbf{2.05 (3.36)} & \textbf{1.14 (1.86)} \\
\bottomrule
\end{tabular}
\end{table}

\begin{table}[h]
\centering
\caption{Pairwise win rates of SHADE vs.\ baselines ($n\in\{5,10,20\}$, pooled queries).}
\label{tab:pairwise_results}
\small
\begin{tabular}{lcccc}
\toprule
Baseline Estimator & Win ($W$) & Loss ($L$) & $n_{\text{valid}}$ & Win Rate (\%) \\
\midrule
Plugin ($k_{\text{obs}}$) & 3372 & 1176 & 5556 & \textbf{74.1} \\
Good-Turing (GT)         & 3125 & 1351 & 5545 & \textbf{69.8} \\
Generalized GT (GGT)     & 3311 & 1358 & 5556 & \textbf{70.9} \\
U-EigV (Laplacian)       & 3523 & 2014 & 5556 & \textbf{63.6} \\
Hybrid (GT+EigV)         & 3187 & 1825 & 5556 & \textbf{63.6} \\
Hybrid-GGT               & 3363 & 2169 & 5556 & \textbf{60.8} \\
\bottomrule
\end{tabular}
\end{table}

\subsection{Incorrectness detection}

We threshold $\hat{\mathbb{H}}_{\text{SHADE}}$ and report ROC AUC for sequence- and response-level labels on four benchmarks (Table~\ref{tab:auc_merged_005}). At $n{=}5$, SHADE attains the highest mean AUC. At $n\in\{8,10\}$, simpler scores such as plugin entropy or NumSets sometimes achieve higher mean AUC despite worse MAE---detection depends on label noise and separability, not only cardinality fidelity.

\begin{table}[H]
\centering
\caption{Incorrectness detection (AUC). Mean column averages four datasets.}
\label{tab:auc_merged_005}
\resizebox{\textwidth}{!}{
\begin{tabular}{llccccccccc}
\toprule
$n$ & Method & \multicolumn{2}{c}{SQuAD} & \multicolumn{2}{c}{NQ-Open} & \multicolumn{2}{c}{TriviaQA} & \multicolumn{2}{c}{HotpotQA} & \textbf{Mean} \\
 & & $AUC_s$ & $AUC_r$ & $AUC_s$ & $AUC_r$ & $AUC_s$ & $AUC_r$ & $AUC_s$ & $AUC_r$ & \\
\midrule
5 & SHADE (Ours) & \textbf{0.711} & 0.666 & 0.678 & \textbf{0.755} & 0.710 & 0.634 & \textbf{0.698} & \textbf{0.697} & \textbf{0.6935} \\
 & $\widehat{\mathbb{H}}_{\mathrm{Hybrid}}$ & 0.710 & 0.665 & 0.677 & 0.753 & 0.709 & 0.633 & 0.697 & 0.696 & 0.6925 \\
 & $\widehat{|S|}_{\mathrm{Hybrid}}$ & 0.696 & \textbf{0.682} & 0.655 & 0.692 & 0.717 & 0.624 & 0.684 & 0.694 & 0.6805 \\
 & NumSets & 0.710 & 0.632 & 0.659 & 0.742 & 0.733 & 0.595 & 0.667 & 0.678 & 0.6770 \\
 & $\widehat{\mathbb{H}}_{\mathrm{Plugin}}$ & 0.699 & 0.644 & 0.651 & 0.735 & 0.724 & 0.591 & 0.661 & 0.663 & 0.6710 \\
 & KLE & 0.590 & 0.623 & \textbf{0.715} & 0.672 & 0.664 & \textbf{0.741} & 0.625 & 0.598 & 0.6535 \\
 & DSE & 0.611 & 0.571 & 0.628 & 0.657 & \textbf{0.757} & 0.653 & 0.627 & 0.608 & 0.6390 \\
\midrule
8 & $\widehat{|S|}_{\mathrm{Hybrid}}$ & 0.701 & \textbf{0.682} & 0.722 & \textbf{0.686} & \textbf{0.778} & 0.626 & \textbf{0.797} & 0.683 & \textbf{0.7094} \\
 & SHADE (Ours) & \textbf{0.742} & 0.681 & 0.725 & 0.681 & 0.765 & 0.625 & 0.772 & 0.681 & 0.7090 \\
 & $\widehat{\mathbb{H}}_{\mathrm{Hybrid}}$ & 0.741 & 0.680 & 0.726 & 0.679 & 0.764 & 0.624 & 0.771 & 0.681 & 0.7083 \\
 & NumSets & 0.739 & 0.675 & \textbf{0.728} & 0.679 & 0.775 & 0.632 & 0.747 & 0.670 & 0.7056 \\
 & $\widehat{\mathbb{H}}_{\mathrm{Plugin}}$ & 0.727 & 0.674 & 0.714 & 0.684 & 0.776 & 0.635 & 0.746 & 0.667 & 0.7029 \\
 & KLE & 0.618 & 0.635 & 0.588 & 0.658 & 0.679 & \textbf{0.782} & 0.622 & \textbf{0.701} & 0.6604 \\
 & DSE & 0.573 & 0.576 & 0.615 & 0.651 & 0.659 & 0.658 & 0.577 & 0.620 & 0.6161 \\
\midrule
10 & $\widehat{\mathbb{H}}_{\mathrm{Plugin}}$ & 0.731 & \textbf{0.683} & 0.743 & \textbf{0.693} & 0.738 & 0.625 & \textbf{0.784} & 0.642 & \textbf{0.7049} \\
 & NumSets & \textbf{0.735} & \textbf{0.683} & 0.747 & 0.685 & 0.737 & 0.612 & 0.768 & 0.637 & 0.7005 \\
 & SHADE (Ours) & 0.733 & 0.679 & \textbf{0.753} & 0.678 & 0.721 & 0.627 & 0.725 & 0.635 & 0.6939 \\
 & $\widehat{\mathbb{H}}_{\mathrm{Hybrid}}$ & 0.732 & 0.677 & 0.752 & 0.676 & 0.724 & 0.626 & 0.724 & 0.635 & 0.6933 \\
 & $\widehat{|S|}_{\mathrm{Hybrid}}$ & 0.677 & 0.654 & 0.723 & 0.687 & \textbf{0.757} & 0.620 & 0.765 & 0.659 & 0.6928 \\
 & KLE & 0.611 & 0.644 & 0.665 & 0.687 & 0.691 & \textbf{0.751} & 0.638 & \textbf{0.680} & 0.6709 \\
 & DSE & 0.603 & 0.591 & 0.604 & 0.659 & 0.726 & 0.718 & 0.645 & 0.646 & 0.6490 \\
\bottomrule
\end{tabular}
}
\end{table}

\section{Conclusion}

SHADE fuses Generalized Good--Turing coverage with the heat-kernel trace of a semantic graph and a finite-sample correction, yielding a single entropy-based risk score. Gains are largest at the smallest sampling budgets. Appendix~\ref{app:exp} lists supplementary protocol notes and societal considerations.

\bibliographystyle{plain}
{\small
\bibliography{main}
}

\appendix

\section{Notation and white-box semantic entropy}
\label{app:prelim}

With access to model probabilities $p(d\mid q,\theta)$, semantic entropy integrates $-\sum_s p(s\mid q,\theta)\log p(s\mid q,\theta)$ over latent classes $s$ \cite{DBLP:conf/iclr/KuhnGF23,Farquhar2024DetectingHI}. Black-box DSE replaces those probabilities with empirical frequencies, which we discussed in Sec.~2.

\section{Experimental protocol}
\label{app:exp}

\paragraph{Supervision.}
Binary labels indicate response-level or sequence-level incorrectness relative to dataset references ($AUC_r$, $AUC_s$), following common practice \cite{Farquhar2024DetectingHI,chen2024quantifying}.

\paragraph{Pseudo-oracle.}
The $N{=}100$ reference is an operational proxy for semantic cardinality; varying $N$ may shift MAE magnitudes while often preserving relative rankings.

\paragraph{Reproducibility.}
Hyperparameters $(\beta,\alpha,\tau)$ are selected once on development data and held fixed; NLI and decoding protocols are shared across benchmarks; random seeds are fixed where applicable. Code and configuration will be released.

\section{Societal considerations}
\label{app:discuss}

Uncertainty scores can support abstention or human review in sensitive domains; poorly calibrated thresholds or clustering failures may create false assurance. We encourage task-specific validation and oversight alongside SHADE.

\end{document}